\pdfoutput=1

\documentclass[11pt]{article}

\usepackage[]{acl}

\usepackage{times}
\usepackage{latexsym}
\usepackage{multirow}
\usepackage{algorithmic}
\usepackage[linesnumbered,ruled,vlined]{algorithm2e}
\usepackage{amssymb}
\usepackage{graphicx}
\usepackage{tabularx,booktabs}
\usepackage{multicol}

\SetKwProg{Init}{init}{}{}

\usepackage[T1]{fontenc}

\usepackage[utf8]{inputenc}

\usepackage{microtype}

\title{Self-Training with Purpose Preserving Augmentation Improves Few-shot Generative Dialogue State Tracking}

\author{Jihyun Lee$^1$, Chaebin Lee$^1$, Yunsu Kim$^{1,2}$, Gary Geunbae Lee$^{1,2}$ \\
  $^1$Graduate School of Artificial Intelligence, POSTECH, Republic of Korea\\
  $^2$Department of Computer Science and Engineering, POSTECH, Republic of Korea\\
  \texttt{\{jihyunlee, leecbin0911, yunsu.kim, gblee\}@postech.ac.kr} \\
}

\begin{document}
\maketitle
\begin{abstract}
In dialogue state tracking (DST), labeling the dataset involves considerable human labor. We propose a new self-training framework for few-shot generative DST that utilize unlabeled data. Our self-training method iteratively improves the model by pseudo labeling and employs \textbf{P}urpose \textbf{P}reserving \textbf{aug}mentation (\textbf{PPaug}) to prevent overfitting. We increase the few-shot (10\%) performance by approximately 4\% on MultiWOZ 2.1 \cite{multiwoz} and enhances the slot-recall 8.34\% for unseen values compared to baseline.

\end{abstract}
\section{Introduction}
\begin{figure}
    \centering
    \includegraphics[width=210pt]{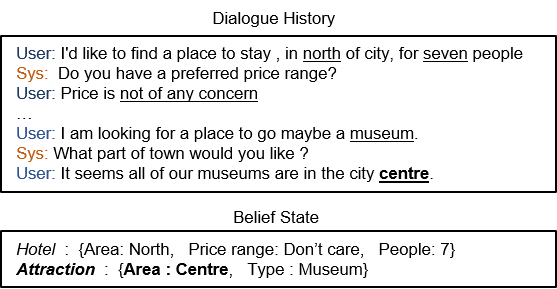}
    \caption{Dialogue example of DST dataset and its belief state. The underlined part of the dialogue is the value of the belief state and has specific information about the user's purpose.}
    \label{fig:dialogue}
\end{figure}

A task-oriented dialogue (TOD) system is a dialogue agent that aims to achieve users' specific purposes which contain all of user's requirements. A TOD system usually consists of several modules, among which dialogue state tracking (DST) is the primary module as it extracts a belief state that includes the user's purpose \cite{young2013pomdp}. A belief state is often represented as a set of slot-value pairs. For example, in Figure~\ref{fig:dialogue}, the belief state has a slot \textit{Attraction-area} and value \textit{Center} that are required to achieve the user’s purpose  \textit{(finding museum in the center of the city)}. Although DST is the core component of the dialogue system, labeling the DST is complex and expensive. Therefore, many few-shot methods have been proposed to address the data scarcity problem \cite{trade, gao2020machine, meta, lin2021leveraging}. 

\begin{figure*}
    \centering
    \includegraphics[width=450pt]{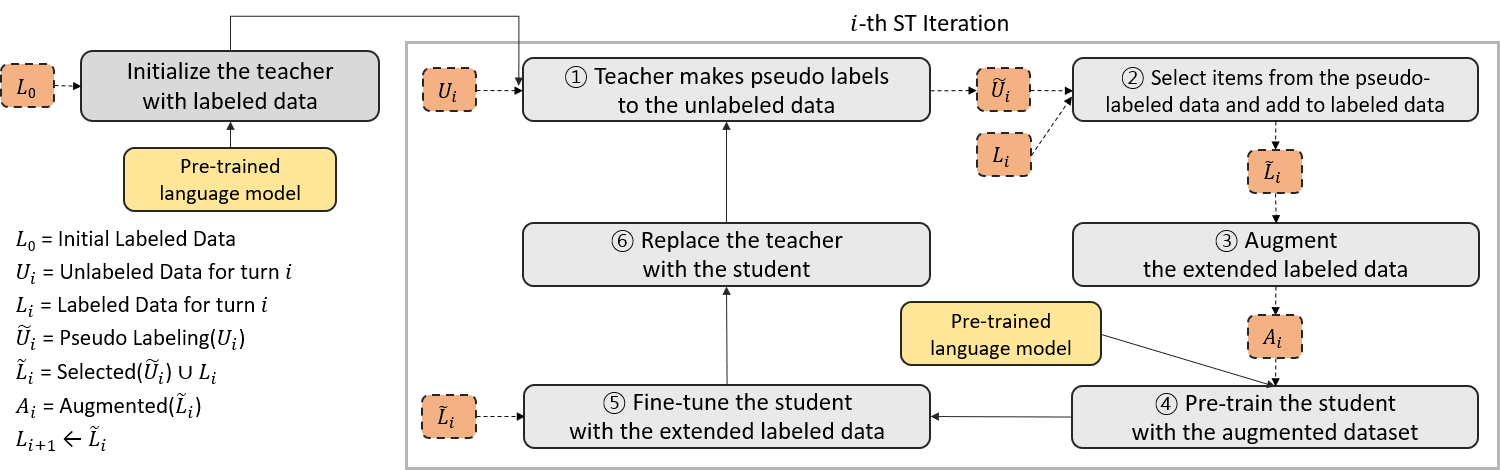}
    \caption{Framework of ST iteration. Each solid line box represents the stage of ST. The dashed small boxes $L$, $U$, $\tilde{U}$, $\tilde{L}$ and $A$ denote labeled, unlabeled, pseudo-labeled, union of labeled and pseudo-labeled, and augmented data, respectively. The solid arrows refer to model update processes and the dashed arrows represent the data flow.}
    \label{fig:loop}
\end{figure*}

In this study, we propose a new few-shot framework that effectively utilizes unlabeled data. In general, the amount of unlabeled dialogue data is abundant compared to labeled data. We focus on this and adopt self-training (ST) \cite{scudder1965probability} as a training strategy for few-shot generative DST. Recently, \citet{mifei} applied ST to few-shot DST to improve the accuracy. They use the classification-based DST as a backbone model and predict values by classifying a pre-defined value set called the ontology. However, their dependence on ontology has a flaw for few-shot DST: because the model can only predict values in the ontology, all plausible values should be collected to train the model. Collecting the ontology requires substantial labor and expert knowledge and is unsuitable for a few-shot DST that aims to reduce human effort. To perform DST without ontology data, we investigate the ST method based on the generative DST model, which is more challenging but ontology-free and can predict unseen values.

Many studies have demonstrated that including augmented data in ST helps prevent overfitting and achieves higher accuracy \cite{laine2016temporal, du2020self, xie2020self}. However, data augmentation in DST is challenging because an augmented sentence should include the belief state information needed to achieve the user's purpose. There have been some attempts to augment DST data \cite{coco, mifei}, but these approaches need to fine-tune a task-specific model for augmentation. To simplify the augmentation process, we developed a new mask infilling method that does not require fine-tuning. We integrate this method into our ST framework and name it \textbf{P}urpose-\textbf{P}reserving \textbf{aug}mentation \textbf{PPaug}.

\section{Method}

\subsection{Dialogue State Tracking (DST)}
\label{problem_statement}
DST extracts the belief states from the user and the system's conversation history. The conversation history for a turn $t$ is denoted as $C_t$ = ($x_1$, $y_2$, $x_2$, $y_2$ ..., $x_t$ ) where $x_i$ is a user utterance and $y_i$ is a system utterance. A belief state has the information required to achieve user's purpose and is denoted as $B_t$ for turn $t$. A belief state consists of slot–value pairs ($s$, $v$).
Given dataset $C_t$ and prompt $m$, we use negative log-likelihood as a loss function as 

\begin{equation}
  L = -\sum_{i=1}^{t} \log p(B_i|m;C_{i})\textnormal{,}
  \label{belief_loss}
\end{equation}

where $m$ is \textit{"translate dialogue to belief state :"}\cite{pptod}.

\subsection{Self-Training (ST)}
\label{self-training}
\subsubsection{Initialization}
The ST method utilizes both labeled data $L$ as well as unlabeled data $U$ in training. Before starting an ST iteration, we trained a teacher model with initial labeled data $L_0$, which contain only true labeled data, not pseudo-labeled data. The student and teacher models have the same model size.

\subsubsection{Pseudo Labeling and Selection}
At the beginning of the \textit{i}-th ST iteration (Figure~\ref{fig:loop}), the teacher model makes pseudo-labeled data $\tilde{U}_i$ by pseudo-labeling unlabeled data $U_i$ and calculates the confidence score for each pseudo label. The softmax value is generally used as the confidence score for the classification model \cite{bank2018improved, he2019revisiting, zou2019confidence}. However, the generative model does not have an explicit softmax value for the predicted pseudo label. Therefore, we newly propose using the average softmax value as a confidence score for generative DST. Then, the selection module selects top-$k$\% items using the confidence score. We move the selected items to the current labeled data $L_i$ from $\tilde{U}_i$ and denote the extended data as $\tilde{L}_i$. For hyperparameter $k$\%, we experimentally use 50\%. The confidence score $S$ is as follows
\begin{equation}
    \textit{confidence score} = \frac{1}{N}\sum_{i=1}^{N}\frac{e^{w_i}}{\sum_{j\in \mathbb{V}}e^{w_j}}
\end{equation}
where $N$, $w_i$ and $\mathbb{V}$ means the length of the word sequences, $i$-th word embedding and vocabulary set, respectively.

\subsubsection{Student Training}
\label{student_training}

After the pseudo labeling and selection, we create an augmented dataset $A_i$ using $\tilde{L}_i$ and train the student model with $A_i$ and $\tilde{L}_i$. In the machine translation domain, \citet{he2019revisiting} used noised data in the pre-training stage and achieved favorable regularization for the subsequent fine-tuning. Following their success, we first pre-trained the model with $A_{i}$ as noise and then fine-tuned it with the $\tilde{L}_i$. The best student model for validation data in the fine-tuning stage becomes a teacher model in next iteration, and the extended data $\tilde{L}_i$ is used as labeled data $L_{i+1}$. We also set $U_{i+1}$ as $U_i$ after deleting selected top-$k$\% items.

\subsection{Purpose-Preserving Data Augmentation}
\label{Augmentation}

\begin{figure}
    \centering
    \includegraphics[width=205pt]{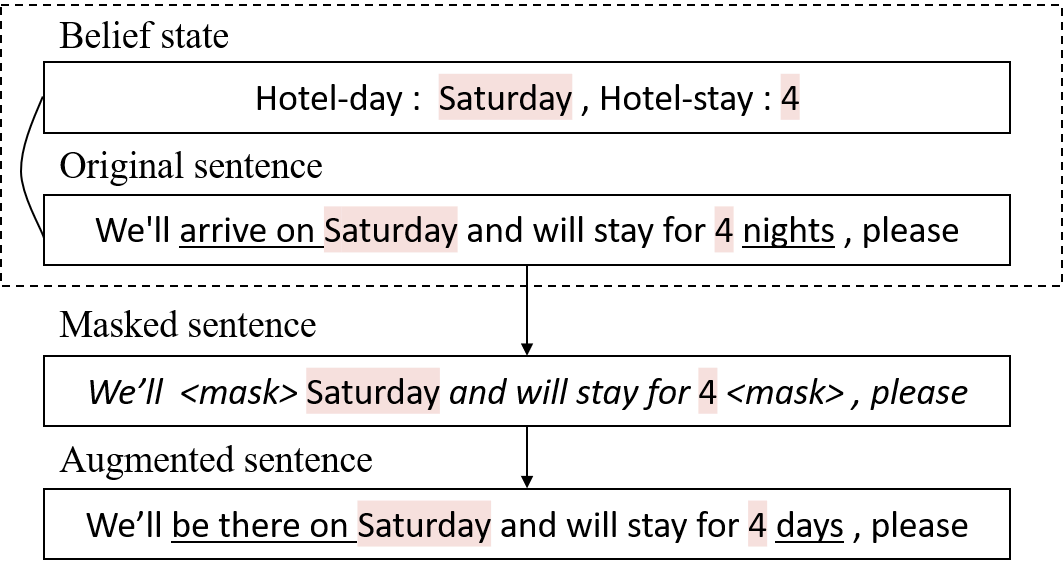}
    \caption{An illustrative example of PPaug. PPaug maintains tokens overlapped with the belief state.}
    \label{fig:aug}
\end{figure}

To prevent overfitting and get more accurate pseudo label \cite{laine2016temporal, du2020self, xie2020self}, we conduct data augmentation in student training. For data augmentation, we leverage the masked language model (MLM). The MLM augments a sentence by replacing the tokens with \textit{<mask>} and infilling it with appropriate tokens. For augmented sentences in DST, containing the belief state information is critical to accurately perform the user's conversation purpose. To do so, when choosing the tokens to be masked, we exclude the tokens overlapped with the belief state, which has important information about the user's purpose. As we change less-important tokens (not overlapped with belief state), our method does not need task-specific fine-tuning. Our method, called purpose-preserving augmentation PPaug, is described in Figure~\ref{fig:aug}. 

PPaug has certain advantages when combined with ST. During each ST iteration, the teacher model assigns a pseudo-label to unlabeled data. When augmenting the data, we utilize not only the initial gold label, but also the pseudo label. This enables a more diverse augmented dataset as the pseudo label increases through each ST iteration.

\section{Experiments}
\subsection{Setup}
To examine our ST and augmentation method in a few-shot environment, we conduct an experiment in which only 10\% of data are labeled and the others are unlabeled. We used the MultiWOZ 2.1 dataset \cite{multiwoz}, which is the most frequently used benchmark in TOD research. It has seven domains (\textit{Hotel, Restaurant, Attraction, Train, Taxi, Hospital, and Police}) and includes 8,000 elements of dialogue realted to tour information. For evaluation, we mainly use joint goal accuracy (JGA), which is the number of correct turns divided by the total number of turns; a turn is counted as correct if all of its predicted slots and values match the true slot-value pairs.

We employ the pre-trained T5-small \cite{t5} for our backbone DST structure. We obtain the generated belief state ($B_t$) as an output in a natural language form as in Figure~\ref{fig:dialogue}. For data augmentation, we use RoBERTa-base \cite{roberta} as an MLM model. Appendix~\ref{appendix:main} has more details of the implementation.

\subsection{Experiment of ST with PPaug}
\label{subsec:experiment of ST with PPaug}
% Ablation study

\begin{table}[]
\centering
{\small
\begin{tabular}{ll}
\hline

Model & JGA [\%]\\ \hline
\multicolumn{2}{c}{\textit{Few-shot setting}} \\ \hline
T5                    &40.98 $\pm$ 0.71   \\ \hline
+ PPaug ($L_0$) &41.32 $\pm$ 0.34   \\
+ ST               &42.17 $\pm$ 1.04\\
+ ST + PPaug w/o pseudo label ($L_0$)           &42.75 $\pm$ 0.34 \\
+ ST + PPaug ($\tilde{L}$) &\textbf{44.09 $\pm$ 0.10}   \\ \hline 
\multicolumn{2}{c}{\textit{Full data}} \\ \hline
 T5          & 52.06 $\pm$ 0.53  \\  \hline         

\end{tabular}
}
\caption{Ablation study of our model reporting the joint goal accuracy (JGA) on MultiWOZ 2.1 in a few-shot (10\%) and a full-data setting.}
\label{tab:ablation study}
\end{table}
\noindent
\textbf{Ablation Study} To evaluate and analyze the contribution of each method we applied, we perform the ablation study in a few-shot environment (10\% of the labeled data). We add our methods one by one from the baseline model (T5). 
As summarized in Table~\ref{tab:ablation study}, augmentation (41.32\%) and ST (42.17\%) both increase the baseline accuracy and show more significant improvement (44.09\%) when used together. In addition, we examine the effect of the pseudo label in data augmentation. Data augmentation without a pseudo label (using only the initial gold label $L_0$) has a lower accuracy (42.75\%) than with the pseudo label $\tilde{L}$ (44.09\%). This shows that the teacher model's pseudo label improves the data augmentation quality. For reference, we added JGA in a full data setting.\newline

% In-train and Unseen Value

\begin{table}[]
\centering
{\small
\begin{tabular}{lll}
\hline
\multirow{2}{*}{Model} & \multicolumn{2}{l}{slot-recall {[}\%{]}} \\ 
                       & In-train           & Unseen           \\ \hline
T5                    &89.17 & 38.24            \\
+ PPaug ($L_0$) &88.87 & 39.05                \\
+ ST               &89.00 & 40.78\\
+ ST+PPaug w/o pseudo label ($L_0$)           &88.80 & 42.43 \\
+ ST + PPaug ($\tilde{L}$) &89.27 & \textbf{46.58}\\ \hline 
\end{tabular}
\caption{Analysis for In-train and Unseen Values reporting few-shot (10\%) slot-recall on the MultiWOZ 2.1 test set.}
\label{tab:out-train analysis}
}
\end{table}
\noindent
\textbf{In-train and Unseen Value} Compared to ST for classification DST, our ST method has the advantage of generating values that are not present in the train data. To explore the effect of ST and PPaug on unseen values, we divide test dataset values into in-train (included in train data, about 39,000 values) and unseen(not included in train data, about 1,000 values) and evaluate the slot-recall\footnote{
We referred to https://github.com/jasonwu0731/trade-dst}.

\begin{equation}{\small
    \textnormal{slot-recall} = {\textnormal{\# of slot-value predicted correctly}\over\textnormal{\# of slot-value in the test dataset}}
\label{equ:slot-recall}
}\end{equation}
Our PPaug (39.05\%)  and ST (40.78\%) both helps to generate unseen values (Table~\ref{tab:out-train analysis}). Especially given that ST can use 90\% of unlabeled data, it is more helpful than only augmentation. Furthermore, using both ST and PPaug (46.58\%) shows an 8.34\% improvement over the baseline (38.24\%) in unseen values. However, in-train values show a similar result. Training with 10\% of the labeled dataset is already sufficient for in-train value, so ST and PPaug have little effect.\newline

% Amount of labeled data.

\begin{table}[]
{\small
\begin{tabular}{@{}llllll@{}} \hline
Labeled-data         & 5\% & 10\% & 20\% & 30\% & 40\% \\ \hline
T5           &  34.77   & 40.98  & 44.72    & 46.63 & 47.88     \\
+ ST + PPaug & 39.35 & 44.09 & 47.33 &  48.39 &  48.93  \\ \hline
Increasing Rate & 13.17 & 7.59 & 5.84 & 3.77  & 2.19\\ \hline
\end{tabular}
\caption{JGA [\%] and increasing rate [\%] with respect to the amount of labeled data in a few-shot setting .}
\label{tab:amount of data}
}
\end{table}

\noindent
\textbf{Amount of Labeled Data} To examine the effectiveness of ST and PPaug in diverse few-shot environments, we experiment by changing the amount of labeled data to 5\%, 10\%, 20\%, 30\%, and 40\%. Our ST and PPaug method has a more pronounced effect when there are less labeled data than when there are enough labeled data (Table~\ref{tab:amount of data}). ST strives to utilize unlabeled data, and PPaug aims to supplement the insufficient data. Therefore, when the labeled data is not enough, our method is more helpful in increasing the accuracy

\subsection{Analysis of ST}
\label{subsec:Analysis of ST}
% Confidence score for generative DST.

\begin{table}[]
\centering
\small
\begin{tabular}{ll}
\hline
Selection Criteria & JGA [\%] \\ \hline
Max & 42.67 \\
Random & 43.39 \\ 
\textbf{Average} & \textbf{44.09} \\ \hline
\end{tabular}
\caption{Comparison with other selection criteria (confidence score) in ST. Reporting few-shot (10\%) JGA.}
\label{tab:criteria}
\end{table}

\noindent\textbf{Selection Criteria (confidence score).} In Table~\ref{tab:criteria}, we compare our proposed selection criteria (average softmax value of tokens) with (i) Max: max softmax value of tokens; and (ii) Random: random softmax value of tokens. Our method shows better accuracy than other selection criteria, and 'Max' performs worse than the 'Random' method. There are some tokens that the decoder produces frequently irrespective of the text input (e.g., slot name or punctuation ":"), and the softmax value of these tokens is relatively high than other tokens. Therefore, the max softmax value does not represent the model's confidence of the generated belief state well and even worse than the randomly chosen value.\newline

% selection method

\begin{table}[]
\centering
\small
\begin{tabular}{ll}\hline

Selection Method     & JGA [\%] \\ \hline
Random-$50$\%       & 43.08    \\
Select-All               &     43.56    \\
Top-$50$\% (Proposed approach) &   \textbf{44.09}\\ \hline \hline

Top-$20$\% &           43.41\\ 
Top-$50$\% (Proposed approach) &   \textbf{44.09}\\ 
Top-$80$\% &         43.10 \\     \hline   

\end{tabular}

\caption{Comparison with other selection methods reporting few-shot (10\%) JGA.}

\label{tab:selector}
\end{table}

\noindent\textbf{Selection Methods} In Table~\ref{tab:selector}, we compare our selection method (top-$k$\%) with other methods including random-$k$\% and select-all \cite{du2020self, strata}. These methods show lower accuracy than using top-$k$\%. In addition, we change the hyperparameter $k$ and compare the results. In this experiment, 50\% shows the best accuracy. When using the top-80\% selector, the model is trained with inaccurate pseudo-labeled data in the early training phase, degrading the performance. Top-20\% is slightly better than top-80\% but converges slowly compared to others.\newline

% Training method in the student model

\begin{table}[]
\small
\centering
\begin{tabular}{ll} \hline
Training method & JGA [\%] \\ \hline
Fine-tuning ($A$ + $\tilde{L}$) & 43.58  \\
Pre-training ($A$) + fine-tuning ($\tilde{L}$) & \textbf{44.09} \\ \hline
\end{tabular}
\caption{Comparison of training method in student model training reporting few-shot (10\%) JGA.}
\label{tab:pre-train fine-tune}
\end{table}

\noindent\textbf{Training Method in Student Model} We pre-train the student model using augmented data ($A$), then fine-tune it with extended labeled data ($\tilde{L}$) in order to utilize the regularization effect of $A$. To see the effect of this separation, we train $A$ and $\tilde{L}$ in the same fine-tuning step as in \citet{mifei} and compare the result. In table~\ref{tab:pre-train fine-tune}, the accuracy of the 'pre-train ($A$) and fine-tuning ($\tilde{L}$)' model is better than 'fine-tuning ($A$ + $\tilde{L}$)'. This indicates that separating the training step is better for preventing overfitting and making a more accurate result.

\begin{figure*}
    \centering
    \includegraphics[width=425pt]{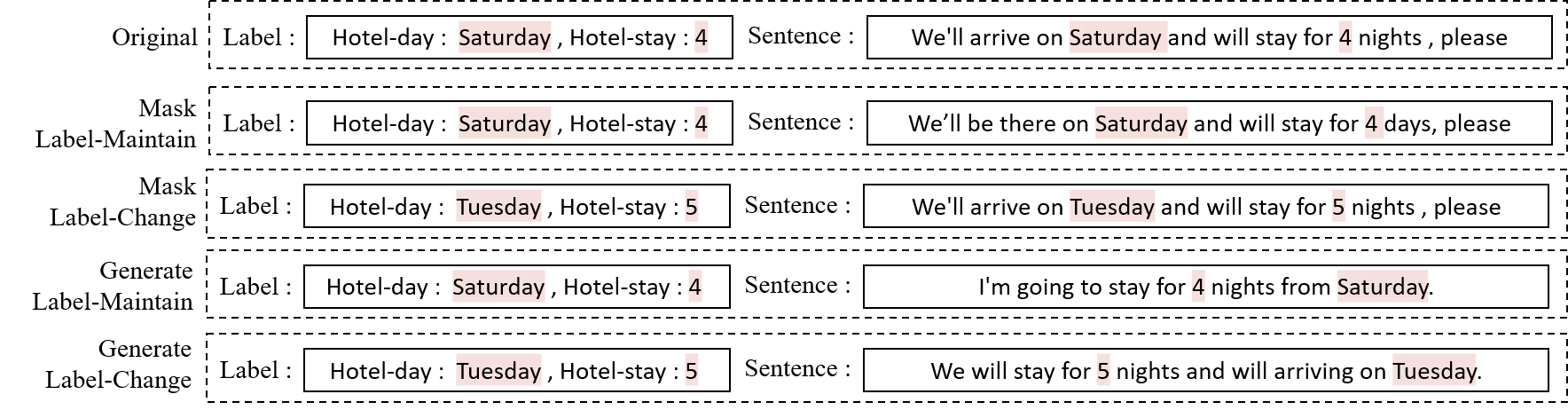}
    \caption{An explanation of variants of PPaug method. The augmented examples are in Appendix~\ref{appendix:example of variants} }
    \label{fig:futuer_example}
\end{figure*}

\subsection{Analysis of PPaug}
\label{subsec:analysis of ppaug}
\begin{table}[]
\centering
\small
\begin{tabular}{ll}\hline
                 &JGA [\%] \\ \hline
T5 + ST &    42.81 \\\hline
+ EDA                 & 42.73    \\
+ AEDA                & 42.97    \\
+ Back translation    & 43.31    \\
+ CoCoAug             & 43.37    \\
+ PPaug   & \textbf{44.09} \\\hline
\end{tabular}

\caption{Comparison with other augmentation methods using the MultiWOZ 2.1 dataset in ST.}
\label{tab:aug_comp}
\end{table}

\noindent\textbf{Other Augmentation Method} Table~\ref{tab:aug_comp} compares our text augmentation method with other commonly used methods, including EDA \cite{eda}, AEDA \cite{aeda}, back-translation \cite{back_translation}, and CoCoAug \cite{li2020coco}.  We apply each augmentation method to the ST framework. The result shows that our PPaug achieves higher accuracy than the other methods. Note that among the experimental results, EDA has the lowest result. This is because EDA randomly drops or changes the words for augmentation, making it difficult to maintain the user's original purpose. This shows the importance of protecting the original user's purpose for augmentation in DST.\newline

\begin{table}[]
\centering
\small
\begin{tabular}{lll}
\hline
Language model                           & Label     & JGA [\%] \\ \hline
\multirow{2}{*}{Mask}           & Maintaining & \textbf{44.09}  \\
                                & Changing    & 42.67 \\\hline
\multirow{2}{*}{Generative}     & Maintaining & 44.02  \\
                                & Changing    & 43.37 \\ \hline
\end{tabular}

\caption{The results of promising variants of PPaug on the MultiWOZ2.1 dataset in ST based on few-shot (10\%) environment.}
\label{tab:promising}
\end{table}

\noindent\textbf{Variants of PPaug} In this experiment, we investigate other variants of PPaug. Each variant is applied to the ST in a few-shot environment (10\% of labeled data is available). We examine an MLM or a generative model (Gen) as the pre-trained model and also distinguish the cases with respect to the label states: maintaining (Maintain) and changing (Change). 
For the generative model, we trained the model as in \citet{coco} with 10\% of the labeled data. Figure~\ref{fig:futuer_example} and Appendix~\ref{Abstract:future} provide illustration and details of the implementation.

The first method (MLM-Maintain) is the same as PPaug; it shows the best result among the set of compared methods (Table~\ref{tab:promising}). 
The second method, MLM-Change, has lower performance than PPaug. Unlike MLM-Maintain, MLM-Change can directly change the user's main purpose (utterance and belief state tokens). The MLM-Change model freely changes the user's purpose to the domain that is not included in MultiWOZ2.1 (Appendix~\ref{appendix:example of variants}). This confuses the model. 
The Gen-Maintain and Gen-Change methods produce sentences with a generative language model. Gen-Maintain and Gen-Change each obtain quite reasonable results, but their performances are lower than PPaug. Note that only 10\% of the labeled dataset is available in this few-shot experiment, and it may not be sufficient for fine-tuning the generative augmentation model. This leads to relatively poor augmented sentences. 
When the labeled data is scarce, MLM-Maintain performs better than the generative models, which need to be fine-tuned.
Appendix~\ref{appendix:example of variants} provides an examples of the augmentations.

\section{Conclusion}
This study proposes ST framework suitable for generative DST and devises a new effective data augmentation method (PPaug). In the ablation study, not only does the proposed ST and PPaug individually improve the accuracy, but they show a synergistic effect when operated together. Additionally, compared to the baseline model, the performance in generating unseen value is greatly improved. As this is the first attempt to adopt ST in generative DST, we thoroughly examine the proposed selection process (average of softmax) and training methods in ST analysis. For future work, we experiment with variants of PPaug and discuss their results.
\bibliography{all.bib}

\begin{thebibliography}{27}
\expandafter\ifx\csname natexlab\endcsname\relax\def\natexlab#1{#1}\fi

\bibitem[{Bank et~al.(2018)Bank, Greenfeld, and Hyams}]{bank2018improved}
Dor Bank, Daniel Greenfeld, and Gal Hyams. 2018.
\newblock Improved training for self training by confidence assessments.
\newblock In \emph{Science and Information Conference}, pages 163--173.
  Springer.

\bibitem[{Devlin et~al.(2018)Devlin, Chang, Lee, and Toutanova}]{bert}
Jacob Devlin, Ming-Wei Chang, Kenton Lee, and Kristina Toutanova. 2018.
\newblock Bert: Pre-training of deep bidirectional transformers for language
  understanding.
\newblock \emph{arXiv preprint arXiv:1810.04805}.

\bibitem[{Dingliwal et~al.(2021)Dingliwal, Gao, Agarwal, Lin, Chung, and
  Hakkani-Tur}]{meta}
Saket Dingliwal, Bill Gao, Sanchit Agarwal, Chien-Wei Lin, Tagyoung Chung, and
  Dilek Hakkani-Tur. 2021.
\newblock Few shot dialogue state tracking using meta-learning.
\newblock \emph{arXiv preprint arXiv:2101.06779}.

\bibitem[{Du et~al.(2020)Du, Grave, Gunel, Chaudhary, Celebi, Auli, Stoyanov,
  and Conneau}]{du2020self}
Jingfei Du, Edouard Grave, Beliz Gunel, Vishrav Chaudhary, Onur Celebi, Michael
  Auli, Ves Stoyanov, and Alexis Conneau. 2020.
\newblock Self-training improves pre-training for natural language
  understanding.
\newblock \emph{arXiv preprint arXiv:2010.02194}.

\bibitem[{Eric et~al.(2019)Eric, Goel, Paul, Kumar, Sethi, Ku, Goyal, Agarwal,
  Gao, and Hakkani-Tur}]{multiwoz}
Mihail Eric, Rahul Goel, Shachi Paul, Adarsh Kumar, Abhishek Sethi, Peter Ku,
  Anuj~Kumar Goyal, Sanchit Agarwal, Shuyang Gao, and Dilek Hakkani-Tur. 2019.
\newblock Multiwoz 2.1: A consolidated multi-domain dialogue dataset with state
  corrections and state tracking baselines.
\newblock \emph{arXiv preprint arXiv:1907.01669}.

\bibitem[{Gao et~al.(2020)Gao, Agarwal, Chung, Jin, and
  Hakkani-Tur}]{gao2020machine}
Shuyang Gao, Sanchit Agarwal, Tagyoung Chung, Di~Jin, and Dilek Hakkani-Tur.
  2020.
\newblock From machine reading comprehension to dialogue state tracking:
  Bridging the gap.
\newblock \emph{arXiv preprint arXiv:2004.05827}.

\bibitem[{Graves(2012)}]{beam}
Alex Graves. 2012.
\newblock Sequence transduction with recurrent neural networks.
\newblock \emph{arXiv preprint arXiv:1211.3711}.

\bibitem[{He et~al.(2019)He, Gu, Shen, and Ranzato}]{he2019revisiting}
Junxian He, Jiatao Gu, Jiajun Shen, and Marc'Aurelio Ranzato. 2019.
\newblock Revisiting self-training for neural sequence generation.
\newblock \emph{arXiv preprint arXiv:1909.13788}.

\bibitem[{Karimi et~al.(2021)Karimi, Rossi, and Prati}]{aeda}
Akbar Karimi, Leonardo Rossi, and Andrea Prati. 2021.
\newblock Aeda: An easier data augmentation technique for text classification.
\newblock \emph{arXiv preprint arXiv:2108.13230}.

\bibitem[{Laine and Aila(2016)}]{laine2016temporal}
Samuli Laine and Timo Aila. 2016.
\newblock Temporal ensembling for semi-supervised learning.
\newblock \emph{arXiv preprint arXiv:1610.02242}.

\bibitem[{Li et~al.(2020{\natexlab{a}})Li, Yavuz, Hashimoto, Li, Niu, Rajani,
  Yan, Zhou, and Xiong}]{coco}
Shiyang Li, Semih Yavuz, Kazuma Hashimoto, Jia Li, Tong Niu, Nazneen Rajani,
  Xifeng Yan, Yingbo Zhou, and Caiming Xiong. 2020{\natexlab{a}}.
\newblock Coco: Controllable counterfactuals for evaluating dialogue state
  trackers.
\newblock \emph{arXiv preprint arXiv:2010.12850}.

\bibitem[{Li et~al.(2020{\natexlab{b}})Li, Yavuz, Hashimoto, Li, Niu, Rajani,
  Yan, Zhou, and Xiong}]{li2020coco}
Shiyang Li, Semih Yavuz, Kazuma Hashimoto, Jia Li, Tong Niu, Nazneen Rajani,
  Xifeng Yan, Yingbo Zhou, and Caiming Xiong. 2020{\natexlab{b}}.
\newblock Coco: Controllable counterfactuals for evaluating dialogue state
  trackers.
\newblock \emph{arXiv preprint arXiv:2010.12850}.

\bibitem[{Lin et~al.(2021)Lin, Liu, Moon, Crook, Zhou, Wang, Yu, Madotto, Cho,
  and Subba}]{lin2021leveraging}
Zhaojiang Lin, Bing Liu, Seungwhan Moon, Paul Crook, Zhenpeng Zhou, Zhiguang
  Wang, Zhou Yu, Andrea Madotto, Eunjoon Cho, and Rajen Subba. 2021.
\newblock Leveraging slot descriptions for zero-shot cross-domain dialogue
  state tracking.
\newblock \emph{arXiv preprint arXiv:2105.04222}.

\bibitem[{Liu et~al.(2019)Liu, Ott, Goyal, Du, Joshi, Chen, Levy, Lewis,
  Zettlemoyer, and Stoyanov}]{roberta}
Yinhan Liu, Myle Ott, Naman Goyal, Jingfei Du, Mandar Joshi, Danqi Chen, Omer
  Levy, Mike Lewis, Luke Zettlemoyer, and Veselin Stoyanov. 2019.
\newblock Roberta: A robustly optimized bert pretraining approach.
\newblock \emph{arXiv preprint arXiv:1907.11692}.

\bibitem[{Loshchilov and Hutter(2017)}]{adamw}
Ilya Loshchilov and Frank Hutter. 2017.
\newblock Decoupled weight decay regularization.
\newblock \emph{arXiv preprint arXiv:1711.05101}.

\bibitem[{Mi et~al.(2021)Mi, Zhou, Cai, Kong, Huang, and Faltings}]{mifei}
Fei Mi, Wanhao Zhou, Fengyu Cai, Lingjing Kong, Minlie Huang, and Boi Faltings.
  2021.
\newblock Self-training improves pre-training for few-shot learning in
  task-oriented dialog systems.
\newblock \emph{arXiv preprint arXiv:2108.12589}.

\bibitem[{Raffel et~al.(2020)Raffel, Shazeer, Roberts, Lee, Narang, Matena,
  Zhou, Li, Liu et~al.}]{t5}
Colin Raffel, Noam Shazeer, Adam Roberts, Katherine Lee, Sharan Narang, Michael
  Matena, Yanqi Zhou, Wei Li, Peter~J Liu, et~al. 2020.
\newblock Exploring the limits of transfer learning with a unified text-to-text
  transformer.
\newblock \emph{J. Mach. Learn. Res.}, 21(140):1--67.

\bibitem[{Scudder(1965)}]{scudder1965probability}
Henry Scudder. 1965.
\newblock Probability of error of some adaptive pattern-recognition machines.
\newblock \emph{IEEE Transactions on Information Theory}, 11(3):363--371.

\bibitem[{Sennrich et~al.(2015)Sennrich, Haddow, and Birch}]{back_translation}
Rico Sennrich, Barry Haddow, and Alexandra Birch. 2015.
\newblock Improving neural machine translation models with monolingual data.
\newblock \emph{arXiv preprint arXiv:1511.06709}.

\bibitem[{Su et~al.(2021)Su, Shu, Mansimov, Gupta, Cai, Lai, and Zhang}]{pptod}
Yixuan Su, Lei Shu, Elman Mansimov, Arshit Gupta, Deng Cai, Yi-An Lai, and
  Yi~Zhang. 2021.
\newblock Multi-task pre-training for plug-and-play task-oriented dialogue
  system.
\newblock \emph{arXiv preprint arXiv:2109.14739}.

\bibitem[{Vu et~al.(2021)Vu, Luong, Le, Simon, and Iyyer}]{strata}
Tu~Vu, Minh-Thang Luong, Quoc~V Le, Grady Simon, and Mohit Iyyer. 2021.
\newblock Strata: Self-training with task augmentation for better few-shot
  learning.
\newblock \emph{arXiv preprint arXiv:2109.06270}.

\bibitem[{Wei and Zou(2019)}]{eda}
Jason Wei and Kai Zou. 2019.
\newblock Eda: Easy data augmentation techniques for boosting performance on
  text classification tasks.
\newblock \emph{arXiv preprint arXiv:1901.11196}.

\bibitem[{Wolf et~al.(2020)Wolf, Debut, Sanh, Chaumond, Delangue, Moi, Cistac,
  Rault, Louf, Funtowicz, Davison, Shleifer, von Platen, Ma, Jernite, Plu, Xu,
  Scao, Gugger, Drame, Lhoest, and Rush}]{wolf2020huggingfaces}
Thomas Wolf, Lysandre Debut, Victor Sanh, Julien Chaumond, Clement Delangue,
  Anthony Moi, Pierric Cistac, Tim Rault, Rémi Louf, Morgan Funtowicz, Joe
  Davison, Sam Shleifer, Patrick von Platen, Clara Ma, Yacine Jernite, Julien
  Plu, Canwen Xu, Teven~Le Scao, Sylvain Gugger, Mariama Drame, Quentin Lhoest,
  and Alexander~M. Rush. 2020.
\newblock \href {http://arxiv.org/abs/1910.03771} {Huggingface's transformers:
  State-of-the-art natural language processing}.

\bibitem[{Wu et~al.(2019)Wu, Madotto, Hosseini-Asl, Xiong, Socher, and
  Fung}]{trade}
Chien-Sheng Wu, Andrea Madotto, Ehsan Hosseini-Asl, Caiming Xiong, Richard
  Socher, and Pascale Fung. 2019.
\newblock Transferable multi-domain state generator for task-oriented dialogue
  systems.
\newblock \emph{arXiv preprint arXiv:1905.08743}.

\bibitem[{Xie et~al.(2020)Xie, Luong, Hovy, and Le}]{xie2020self}
Qizhe Xie, Minh-Thang Luong, Eduard Hovy, and Quoc~V Le. 2020.
\newblock Self-training with noisy student improves imagenet classification.
\newblock In \emph{Proceedings of the IEEE/CVF conference on computer vision
  and pattern recognition}, pages 10687--10698.

\bibitem[{Young et~al.(2013)Young, Ga{\v{s}}i{\'c}, Thomson, and
  Williams}]{young2013pomdp}
Steve Young, Milica Ga{\v{s}}i{\'c}, Blaise Thomson, and Jason~D Williams.
  2013.
\newblock Pomdp-based statistical spoken dialog systems: A review.
\newblock \emph{Proceedings of the IEEE}, 101(5):1160--1179.

\bibitem[{Zou et~al.(2019)Zou, Yu, Liu, Kumar, and Wang}]{zou2019confidence}
Yang Zou, Zhiding Yu, Xiaofeng Liu, BVK Kumar, and Jinsong Wang. 2019.
\newblock Confidence regularized self-training.
\newblock In \emph{Proceedings of the IEEE/CVF International Conference on
  Computer Vision}, pages 5982--5991.

\end{thebibliography}
\newpage
\quad\newpage
\appendix

\section{Appendix}
\label{appendix:main}
\subsection{Detailed implementation}
In ST, we iterate the training loop until we reach 10 epochs with early stopping. To select the pseudo label, we select the top-50\% in terms of the confidence score. In augmentation, we randomly choose 20\% of tokens to change as \textit{<mask>}, slightly more than the default setting of BERT \cite{bert} (15\%), and double the size of the original sentence set. To pre-train the student model, we train the model for 20 epochs with early stopping, and in fine-tuning, we train for 10 epochs with early stopping. We implement a backbone generative DST model using T5-small \cite{t5}, which has six encoder/decoder layers and a hidden model of size 512. We use an NVIDIA A5000 graphics processing unit for all training, and an AdamW \cite{adamw} optimizer with a learning rate of 5e-5. We set the batch size to 128 and implement T5-small based on the Huggingface Library \cite{wolf2020huggingfaces}.

\subsection{Detailed implementation for Table~\ref{tab:promising} results}
\label{Abstract:future}
Here, we explain the implementation of the experiment in Section \ref{subsec:analysis of ppaug} variants of ppaug. For MLM augmentation, we use the pre-trained RoBERTa-base \cite{roberta}, and for the generative language model (Gen), we use T5-small \cite{t5}. The generative model is fine-tuned to generate conditioned utterances given $C_t$ and $B_t$ using 10\% of the labeled data following \citet{coco}. 

The MLM-Maintain method is the same as our PPaug method described in Section \ref{Augmentation}. For the MLM-Change method, when choosing the tokens to replace \textit{<mask>}, we choose the tokens overlapped with the belief state and also change the belief state's tokens as per the MLM's result. The Gen-Maintain method generates various sentences via beam search \cite{beam} using the same label as the original text. Finally, the Gen-Change method generates various sentences via beam search \cite{beam} with a changed label. In Gen-Change, to change the label appropriately, we build a slot–value dictionary from the 10\% of initial labeled 

\subsection{Masking Rate of PPaug}
\label{subsec:masking}
\begin{table}[]
\centering
\begin{tabular}{ll}
\hline
Masking rate [\%] & JGA [\%] \\
\hline
10 & 43.56\% \\
20 (Proposed approach) & \textbf{44.09}\% \\
40 & 43.53\% \\
60 & 43.15\% \\
80 & 43.22\% \\
\hline
\end{tabular}
\caption{Comparison with each other masking rate in the PPaug reporting few-shot (10\%) JGA on the MultiWOZ 2.1 test set.}
\label{tab:masking rate}
\end{table}

In PPaug, we randomly choose the some tokens to change as \textit{<mask>}. We experiment different masking rate (10\%, 20\%, 40\%, 60\% and 80\%) to find best masking rate . Table \ref{tab:masking rate} summarizes the results of each masking rates in the PPaug.

\onecolumn

\subsection{Example of Augmented Sentence with Variants of PPaug}
\label{appendix:example of variants}
This section show the examples of Variants of PPaug and its error type. \\

\noindent\textbf{Example of MLM-Change} 

\begin{table}[h]
\small
\begin{tabular}{ll} 

\multicolumn{2}{c}{Example 1} \\ \hline
\textit{Original} & \vline
~   i would just like to arrive by \textbf{16:00} please. \\ 
\textit{Belief state} & \vline ~ [train][time] \textbf{16:00} \\ \hline
\textit{Augmented} & \vline ~  i would just like to arrive by \textbf{16:45} please.\\ 
\textit{Belief state} & \vline ~ [train][time] \textbf{16:45} \\ \hline \\

\multicolumn{2}{c}{Example 2 - Replace with out of domain (MultiWOZ2.1) value} \\ \hline
\textit{Original} & \vline
~  is the gonville a good quality \textbf{hotel}? \\ 
\textit{Belief state} & \vline ~ [hotel][value type] \textbf{hotel} \\ \hline
\textit{Augmented} & \vline ~  is the gonville a good quality \textbf{watch} ?\\ 
\textit{Belief state} & \vline ~ [hotel][value type] \textbf{watch} \\ \hline \\

\multicolumn{2}{c}{Example 3 - Replace with inadequate value for slot}\\ \hline
\textit{Original} & \vline ~ i would, thanks. i need a table for 6 on \textbf{sunday}.\\ \textit{Belief state} & \vline ~ [restaurant][value people] 6, [value day] \textbf{sunday} \\ \hline
\textit{Augmented} & \vline ~ i would, thanks. i need a table for 6 on the \textbf{table}.\\ 
\textit{Belief state} & \vline ~ [restaurant][value people] 6, [value day] \textbf{table} \\ \hline \\

\multicolumn{2}{c}{Example 4 - Replace with inadequate value for slot}\\ \hline
\textit{Original} & \vline ~ the restaurant should serve asian \textbf{oriental food}. \\ 
\textit{Belief state} & \vline ~ [restaurant][type] \textbf{oriental food} \\ \hline
\textit{Augmented} & \vline ~  the restaurant should serve asian \textbf{orientalis}. \\ 
\textit{Belief state} & \vline ~ [restaurant][type] \textbf{orientalis} \\ \hline
\end{tabular}
\end{table}

\noindent\textbf{Example of Gen-Maintain}

\begin{table}[h]
\small
\begin{tabular}{ll} 
\multicolumn{2}{c}{Example 1} \\ \hline
\textit{Original} & \vline
~   i would like to find a \textbf{museum} to visit.  \\ 
\textit{Belief state} & \vline ~ [attraction][area] \textbf{museum}\\ \hline
\textit{Augmented} & \vline ~  i am looking for a \textbf{museum} to visit.\\ 
\textit{Belief state} & \vline ~ [attraction][area] \textbf{museum}\\ \hline \\

\multicolumn{2}{c}{Example 2 - Omit the information} \\ \hline
\textit{Original} & \vline
~  i would like to book a room for 3 days starting tuesday. there is a total of \textbf{3 people}.  \\ 
\textit{Belief state} & \vline ~ [hotel][value stay] 3 days [value day] tuesday [value people] \textbf{3 people}  \\ \hline
\textit{Augmented} & \vline ~  i would like to book a room for 3 nights starting on tues\\
\textit{Belief state} & \vline ~ [hotel][value stay] 3 days [value day] tuesday [value people] \textbf{3 people} \\ \hline \\ 

\multicolumn{2}{c}{Example 3 - Omit the information} \\ \hline
\textit{Original} & \vline ~ yes, can you find me a \textbf{cheap} place to eat serving chinese food?\\ 
\textit{Belief state} & \vline ~ [restaurant][price] \textbf{cheap}, [type] chinese \\ \hline
\textit{Augmented} & \vline ~ i would like a chinese restaurant.\\ 
\textit{Belief state} & \vline ~ [restaurant][price] \textbf{cheap}, [type] chinese \\ \hline \\

\multicolumn{2}{c}{Example 4 - Generate informative words not in belief state}\\ \hline
\textit{Original} & \vline ~ i need a train departing from \textbf{kings lynn}. \\ 
\textit{Belief state} & \vline ~ [train][area] \textbf{kings lynn} \\ \hline
\textit{Augmented} & \vline ~  yes, i am looking for a train leaving \textbf{kings lynn} on the same day as my \textbf{hotel check in date}. \\ 
\textit{Belief state} & \vline ~ [train][area] \textbf{kings lynn} \\ \hline
\end{tabular}
\end{table}

\newpage
\noindent\textbf{Example of Gen-Change}

\begin{table}[h]
\small
\begin{tabular}{ll} 

\multicolumn{2}{c}{Example 1} \\ \hline
\textit{Original} & \vline
~   i need a \textbf{guest house} in the \textbf{east} please.   \\ 
\textit{Belief state} & \vline ~ [hotel][type] \textbf{guest house} [area]\textbf{east}\\ \hline
\textit{Augmented} & \vline ~  i would like a \textbf{hotel} in the \textbf{west}.\\ 
\textit{Belief state} & \vline ~ [hotel][type] \textbf{hotel} [area]\textbf{west}\\ \hline \\

\multicolumn{2}{c}{Example 2 - Not matched with slot}\\ \hline
\textit{Original} & \vline
~  i also need information on cambridge book and print gallery .   \\ 
\textit{Belief state} & \vline ~ [attraction][area] cambridge book and print gallery \\ \hline
\textit{Augmented} & \vline ~  yes, i would like to book a \textbf{taxi} to go to the cinema.\\ 
\textit{Belief state} & \vline ~ \textbf{[attraction]} cinema  \\ \hline \\ 

\multicolumn{2}{c}{Example 3 - Omit the information}\\ \hline
\textit{Original} & \vline ~ the center of town would be best .\\ 
\textit{Belief state} & \vline ~ [hotel][area] center of town \\ \hline
\textit{Augmented} & \vline ~ yes please book it for 6 people at 15:00 on tuesday.\\ 
\textit{Belief state} & \vline ~ [hotel][value people] 6 people [time] 15:00 [value day] tuesday [area] \textbf{centre} \\ \hline \\

\multicolumn{2}{c}{Example 4 - Omit the information, Generate informative words not in belief state. } \\ \hline
\textit{Original} & \vline ~ i am looking for places to go in the east side of town . do you have any suggestions ? \\ 
\textit{Belief state} & \vline ~ [attraction][area] east side of town \\ \hline
\textit{Augmented} & \vline ~  i would like to leave the hotel on \textbf{wednesday}. \\ 
\textit{Belief state} & \vline ~ [taxi][departure] hotel [attraction][area] \textbf{west} \\ \hline 
\end{tabular}
\end{table}

%어울리지 않는 belief로 대체 되거나 belief state의 정보를 빼먹는다%
%MLM-Change replaces the tokens with inadequate words for task-oriented dialogue. When they irrelatively replace the tokens, the sentence often miss the important information to achieve user's goal. \\

%Gen-Maintain often generates the sentence without using the some belief states (Example 1, Example 2). Also, they often makes the sentence with more information, but the information is not included in the gold label (Example 3). The factors have an effect on degrading the DST performance, because we do not change the gold label (slot-value) of augmented sentence. \\

%Gen-Change generates the sentence based on the replaced slot-value. The augmented sentences seem quite plausible, but they often generate the sentence that does not match the gold label. In Example 1, the replaced belief state is \textit{[attraction] cinema}, but the generated sentence is related to \textit{[taxi]} rather than \textit{[attraction]}. Also, Gen-Change omits some belief states (Example 2 and 3) ,or generates sentences with more information which is not in the replaced belief state (Example 3). These factors make the DST model degraded.

\end{document}